\documentclass[11pt]{article}
\usepackage[letterpaper,margin=1in]{geometry}
\usepackage[T1]{fontenc}
\usepackage[utf8]{inputenc}
\usepackage{newtxtext}
\usepackage{newtxmath}
\usepackage{graphicx}
\usepackage{amsmath}
\usepackage{booktabs}
\usepackage{microtype}
\usepackage{setspace}
\usepackage{hyperref}
\usepackage{enumitem}
\usepackage{caption}
\usepackage{xcolor}
\usepackage{float}
\hypersetup{colorlinks=true,linkcolor=black,citecolor=black,urlcolor=blue,
 pdftitle={Pilot Early, Commit Late: A Real-Options Model of Enterprise AI Adoption under Rapid Technological Progress},
 pdfauthor={Gaurav Tewari (Omega Venture Partners)},
 pdfsubject={Enterprise artificial intelligence deployment planning under uncertainty, real options, organizational learning, and modularity},
 pdfkeywords={artificial intelligence adoption, AI deployment planning, decision-making under uncertainty, real options, experimentation, organizational learning, modular architecture, investment timing},
 pdfcreator={LaTeX}}
\title{\textbf{Pilot Early, Commit Late: A Real-Options Model of Enterprise AI Adoption under Rapid Technological Progress}\\[0.5em]\large Why Faster Frontier Progress Can Delay Full Deployment While Increasing the Value of Experimentation}
\author{Gaurav Tewari\\\small Omega Venture Partners\\\small \href{https://www.omegavp.com/}{www.omegavp.com}}
\date{September 2026}
\begin{document}
\maketitle
\begin{abstract}
Artificial intelligence presents firms with an unusual timing problem. The technology frontier is improving rapidly, implementation is partly irreversible, and organization-specific capabilities are accumulated through action. This paper develops a two-period decision model of AI deployment under uncertainty in which a firm chooses among immediate deployment, a limited pilot, and waiting. Deployment earns current operating value but exposes the firm to architectural obsolescence; waiting preserves the option to adopt after the frontier is observed; a pilot sacrifices current operating value to build organization-specific learning without full commitment. The model yields five central timing results and a sixth comparative result on where learning occurs. First, a mean-preserving increase in frontier uncertainty raises the value of waiting and piloting but leaves immediate deployment unchanged when its payoff is affine in the frontier. Second, faster expected frontier progress can reduce the relative attractiveness of immediate deployment when deployed architecture captures only a limited share of future improvement. Third, a pilot dominates waiting exactly when the expected value of the capability it builds exceeds its cost. Fourth, sufficiently valuable organization-specific learning creates a nonempty region in which ``pilot early, commit late'' is optimal. Fifth, there is a closed-form modularity threshold above which immediate deployment dominates the best outside option. Sixth, production learning and pilot-specific learning affect the timing margin differently. A continuous-time extension recovers the standard result that uncertainty raises the adoption threshold while capability and modularity lower it. The paper separates deploying, experimenting, and waiting, and shows why rapid progress can rationally increase experimentation without justifying irreversible commitment.\end{abstract}

\section{Introduction}
Enterprise artificial intelligence creates a timing problem that standard adoption language obscures. A firm can deploy now, experiment without committing fully, or wait. Those choices are not interchangeable. Immediate deployment earns operating benefits and may create learning-by-doing, but it also embeds current models, vendors, interfaces, data pipelines, and control structures into an architecture that may age quickly. Waiting preserves flexibility, but produces no organization-specific capability. A pilot occupies the middle: it is costly and does not deliver the full operating payoff, yet it can build knowledge that is difficult to acquire by observation alone. The timing problem is becoming more economically important because the technical frontier is moving unusually fast. The 2026 AI Index documents large one-year gains on difficult benchmarks and continuing convergence among leading model providers (Stanford Institute for Human-Centered Artificial Intelligence, 2026). At the same time, firm-level use remains uneven in depth. Recent surveys and administrative evidence show rapid adoption but substantial concentration by firm size, sector, business function, and managerial practice (Yotzov et al., 2026; Bonney et al., 2026). Field experiments demonstrate material productivity gains in some tasks, but also heterogeneous effects and performance losses outside the technology's effective boundary (Brynjolfsson et al., 2025; Dell'Acqua et al., 2026). These facts make the adoption decision neither a simple race nor a simple wait-and-see exercise. Classic real-options theory explains why uncertainty and irreversibility can delay investment (McDonald and Siegel, 1986; Pindyck, 1991; Dixit and Pindyck, 1994). Technology-adoption research applies the same logic to IT platforms and software projects (Benaroch and Kauffman, 1999; Schwartz and Zozaya-Gorostiza, 2003; Fichman, 2004). A parallel literature emphasizes experimentation and learning. Prototypes, R\&D, and staged inquiry can be valuable precisely because uncertainty is high (Thomke, 1998; Erdogmus, 2002; Ross et al., 2018; Crouzet and Eberly, 2026; Caplin, 2026). In the AI setting, recent work finds that uncertainty can suppress irreversible firm investment and increase the option value of waiting (Gandhi et al., 2026; Yue and Zhou, 2025). Existing work does not jointly model waiting, piloting, and deployment when architecture determines how much future frontier progress an early system can capture. This paper models that joint decision in two periods. At date 0, a firm chooses one of three actions. It may deploy and earn current value, while capturing only a fraction of future frontier improvement. It may wait until date 1, observe the future frontier, and deploy only if profitable. Or it may conduct a pilot that builds organization-specific capability before the same date-1 deployment decision. The model abstracts from implementation detail to isolate commitment, experimentation, and inaction. The analysis produces six results. First, the option payoffs from waiting and piloting are convex in the future frontier. Holding the mean fixed, greater uncertainty weakly raises both values. Immediate deployment is different: when its payoff is affine in the future frontier, the value of deployment depends on the mean but not the dispersion. Uncertainty therefore changes the ranking of actions even when it does not change expected technical progress. Second, faster expected frontier progress can delay commitment. Under a location shift in the future frontier, the value of waiting or piloting rises with the probability that future adoption will be profitable. The value of immediate deployment rises only with the fraction of frontier improvement that the installed architecture can capture. If that fraction is low, better expected future models can make today's full deployment relatively less attractive. This is the paper's AI waiting paradox: technical progress raises the value of AI while increasing the rational incentive to defer irreversible implementation. Third, a pilot is not justified merely because uncertainty is high. It dominates pure waiting only when its organization-specific learning value exceeds its cost. The model yields an exact threshold for that comparison and shows that the threshold increases with the learning produced by the pilot and with first-order improvements in the future frontier. Fourth, ``pilot early, commit late'' defines a distinct policy region. For sufficiently valuable learning and a bounded pilot cost, piloting strictly dominates both immediate deployment and passive waiting. The firm acts now, but the action is designed to preserve the later deployment option rather than exercise it. Fifth, architecture has a precise economic role. A modular deployment captures a larger share of future frontier improvement. The model yields a closed-form modularity threshold above which immediate deployment beats the best of piloting and waiting. Economically, modularity reduces the obsolescence penalty embedded in early commitment. Sixth, the source of learning matters. Learning-by-doing from production raises the value of deploying now, while pilot-specific learning raises the value of staged experimentation. Recommendations such as ``start small'' are incomplete unless they specify whether a small start creates capability that transfers to a scaled system. The paper's contribution is to model three mechanisms jointly: deploy/pilot/wait choice, organization-specific capability accumulation, and architecture-dependent capture of a moving AI frontier. Figure 1 summarizes the decision environment. The rest of the paper proceeds as follows. Section 2 relates the model to real options, IT adoption, experimentation, and recent AI evidence. Section 3 presents the two-period model. Section 4 derives the analytical results. Section 5 provides numerical experiments. Section 6 develops a continuous-time extension. Section 7 discusses implications. Section 8 states the limitations and empirical agenda. Proofs and numerical details appear in the appendices.

\begin{figure}[H]
\centering
\includegraphics[width=0.92\textwidth]{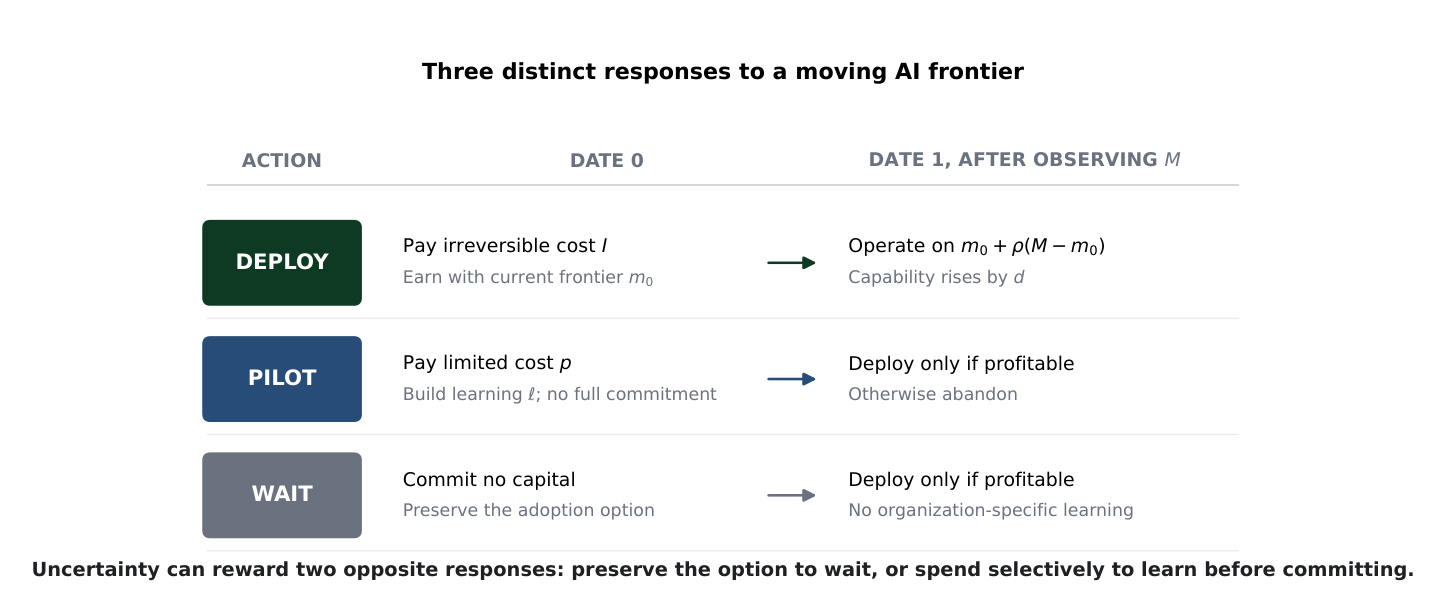}
\caption{Three responses to a moving AI frontier. Immediate deployment combines current operating value with architectural exposure to future progress. A pilot buys organization-specific learning while preserving the later deployment option. Waiting preserves flexibility but creates no internal capability.}
\end{figure}

\section{Related Literature}
\subsection{Irreversibility and the option to wait}
The real-options literature establishes that a firm facing irreversible investment and uncertain future payoffs should generally require more than a positive static net present value before committing. The ability to wait is itself an asset (McDonald and Siegel, 1986; Pindyck, 1991; Dixit and Pindyck, 1994). Empirical work finds that uncertainty can reduce the responsiveness of investment when adjustment is costly or irreversible (Bloom et al., 2007). Strategic-management research extends the logic from asset pricing to organizational flexibility, technology positioning, and staged commitment (McGrath, 1997; Trigeorgis and Reuer, 2017). It also cautions that not every sequential decision is a real option: the option interpretation requires a credible right, bounded downside, and a meaningful future exercise decision (Adner and Levinthal, 2004). Waiting has value because the firm can observe the future frontier before choosing whether to pay the deployment cost. Piloting has value only because it changes the capability available at that later exercise decision. Immediate deployment is irreversible because its implementation cost is sunk.

\subsection{Technology adoption and obsolescence}
Technology investments are especially suited to real-options analysis because implementation costs are partly sunk while technical capability, cost, and standards evolve. Grenadier and Weiss (1997) model investment in a sequence of technological innovations and show how learning from current adoption can affect later choices. Benaroch and Kauffman (1999) and Schwartz and Zozaya-Gorostiza (2003) develop valuation approaches for IT acquisition and development projects under uncertainty. Fichman (2004) synthesizes technology strategy, organizational learning, bandwagons, and adaptation into an option-value framework for IT platforms. The present model focuses on a narrower but increasingly salient feature: an early enterprise AI deployment may continue operating while capturing only part of subsequent frontier improvement. The parameter \(\rho\) makes this exposure explicit. A deployment with low \(\rho\) is not necessarily technically frozen, but upgrades require enough reengineering, vendor change, workflow redesign, or control remediation that the installed system captures little of the external advance. A highly modular system captures more.

\subsection{Experimentation and organization-specific learning}
Experimentation can be an investment rather than a diluted version of deployment. Thomke (1998) shows how prototyping and simulation alter the economics of product-development learning. Erdogmus (2002) values a software prototype as a staged option preceding full development. Ross et al. (2018) document that uncertainty can increase R\&D when firms possess favorable learning conditions. Crouzet and Eberly (2026) formalize the opposing effects of uncertainty on irreversible capital and information-producing R\&D. Caplin (2026) treats frontier investment as staged discovery that builds a reusable stock of planning capital. Related work in entrepreneurship emphasizes sequential financing as a mechanism for experimentation under fundamental uncertainty (Nanda and Rhodes-Kropf, 2016). The pilot in this paper builds capability rather than merely producing a noisy signal. Capability includes the firm's ability to select tasks, prepare data, redesign processes, evaluate outputs, train users, integrate systems, and govern exceptions. This reduced-form treatment is consistent with absorptive-capacity theory: external knowledge creates value only when the organization can recognize, assimilate, and apply it (Cohen and Levinthal, 1990). It also reflects evidence that complementary intangible investment is required before general-purpose technologies generate measured productivity (Brynjolfsson et al., 2021).

\begin{table}[H]
\centering
\includegraphics[width=0.88\textwidth]{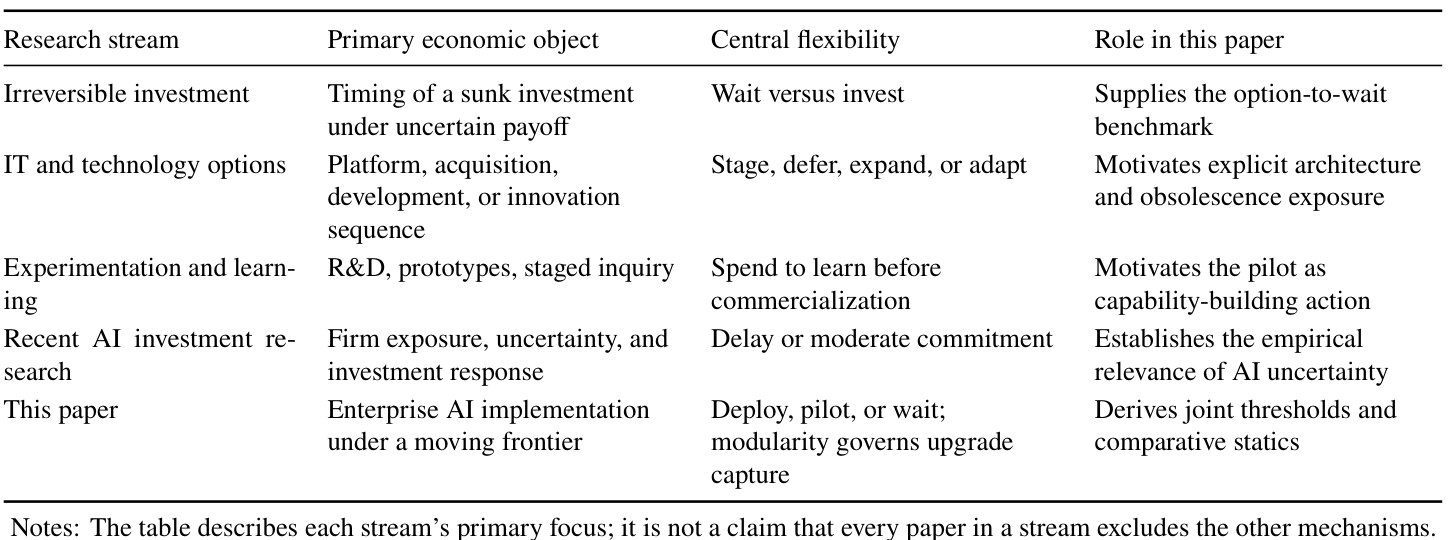}
\caption{Positioning relative to adjacent research.}
\end{table}
\subsection{AI deployment, organizational decision-making, and uncertainty}
Recent AI research treats organizational AI adoption as a deployment and decision problem, not only as a model-performance problem. Lee et al. (2026) document how organizational goals, worker experience, usability, interoperability, and control shape AI integration. Banerjee and Singh (2026) develop an analytical model of human--AI task allocation in organizations, deriving substitution conditions from costs, skills, organizational depth, deployment scale, strategic adaptation, and risk. The present paper addresses a different decision margin: whether an enterprise should deploy, pilot, or wait when the external AI frontier is stochastic and architecture determines how much subsequent progress an installed system can capture.

Recent empirical work makes the timing question concrete. AI use is spreading quickly, but adoption depth remains heterogeneous across firms and functions (Yotzov et al., 2026; Bonney et al., 2026). Task-level studies show both productivity gains and a jagged boundary between tasks on which AI helps and tasks on which it harms performance (Brynjolfsson et al., 2025; Dell'Acqua et al., 2026). Those findings create uncertainty not only about model capability but about organizational fit. Fridgen et al. (2022) apply digital-options thinking to AI adoption in retail banking and emphasize the moving frontier and contextual capabilities. Gandhi et al. (2026) report that firms with greater workforce exposure to generative AI reduced investment following a late-2022 public launch of a widely used generative-AI assistant and interpret the result through parameter uncertainty and the option to wait. Yue and Zhou (2025) examine AI investment, uncertainty exposure, organizational learning, and firm valuation. These studies motivate the present model but do not make piloting, waiting, and modular full deployment simultaneous alternatives in a common enterprise-AI timing problem.

\section{Model}
\subsection{Environment}
There are two dates, \(t=0\) and \(t=1\). The current AI frontier is \(m_0>0\). At date 1 the frontier is a positive random variable \(M\) with finite mean \(\mu_M=\mathbb{E}[M]>m_0\). A higher realization of \(M\) represents a more capable, cheaper, or otherwise more economically useful external AI frontier. The firm has organization-specific capability \(k_0>0\). If the firm operates with capability \(k\) and effective frontier \(m\), gross one-period surplus is

\begin{equation}
\Pi(k,m)=Akm.
\end{equation}
where \(A>0\) scales the economic importance of the use case. Full deployment requires a sunk cost \(I>0\). The discount factor is \(\delta\in(0,1]\). The firm is risk neutral. The model omits financing constraints and strategic competition in order to isolate timing, learning, and modularity. At date 0 the firm chooses among three mutually exclusive actions.

Deploy. The firm pays \(I\), earns current surplus \(Ak_0m_0\), and accumulates learning-by-doing \(d\ge0\). Its installed architecture captures a fraction \(\rho\in[0,1]\) of the improvement between \(m_0\) and \(M\). Date-1 effective frontier is therefore

\begin{equation}
\widetilde{M}(\rho)=m_0+\rho(M-m_0).
\end{equation}
The expected value of deployment is

\begin{equation}
D(\rho)=Ak_0m_0-I+\delta A(k_0+d)\left[m_0+\rho(\mu_M-m_0)\right].
\end{equation}
The parameter \(\rho\) is the paper's measure of modularity or upgrade capture. It can represent model abstraction, vendor portability, separable data and orchestration layers, modular evaluation infrastructure, or an implementation design that avoids embedding a single model's behavior deeply into the workflow.

Wait. The firm makes no date-0 investment. At date 1 it observes \(M\) and deploys only if operating surplus exceeds the cost. Define

\begin{equation}
h_k(m)=[Akm-I]_+.
\end{equation}
\begin{samepage}
The value of waiting is
\begin{equation}
W=\delta\,\mathbb{E}[h_{k_0}(M)].
\end{equation}
\end{samepage}
Because the firm can decline to deploy, \(W\ge0\).

Pilot. The firm pays \(p\ge0\) at date 0 and obtains no full operating surplus. The pilot increases organization-specific capability by \(\ell\ge0\). At date 1 the firm observes \(M\) and deploys only if profitable. Pilot value is

\begin{equation}
P=-p+\delta\,\mathbb{E}[h_{k_0+\ell}(M)].
\end{equation}
A pilot can be abandoned; its downside is bounded by \(p\). It is economically distinct from a miniature production rollout because its modeled payoff is learning, not current operating cash flow. The optimal date-0 policy is

\begin{equation}
V_0=\max\{D(\rho),P,W\}.
\end{equation}
Costless waiting weakly dominates permanent inaction, so a separate no-action choice is unnecessary. A maintenance cost for preserving readiness would create an inaction region without changing the central comparative statics.

\subsection{Interpretation of the parameters}
Three distinctions are important. First, \(M\) is external progress; \(k\) is internal capability. A firm cannot purchase \(k\) simply by waiting for a better model. Second, \(d\) and \(\ell\) capture different learning channels. Full deployment may teach more because it exposes the organization to production conditions, but a pilot can generate useful capability at bounded cost. Third, \(\rho\) does not assert that software literally cannot be upgraded. It measures the economic fraction of the future frontier that the installed system can capture without paying a new sunk implementation cost.

\begin{table}[H]
\centering
\includegraphics[width=0.88\textwidth]{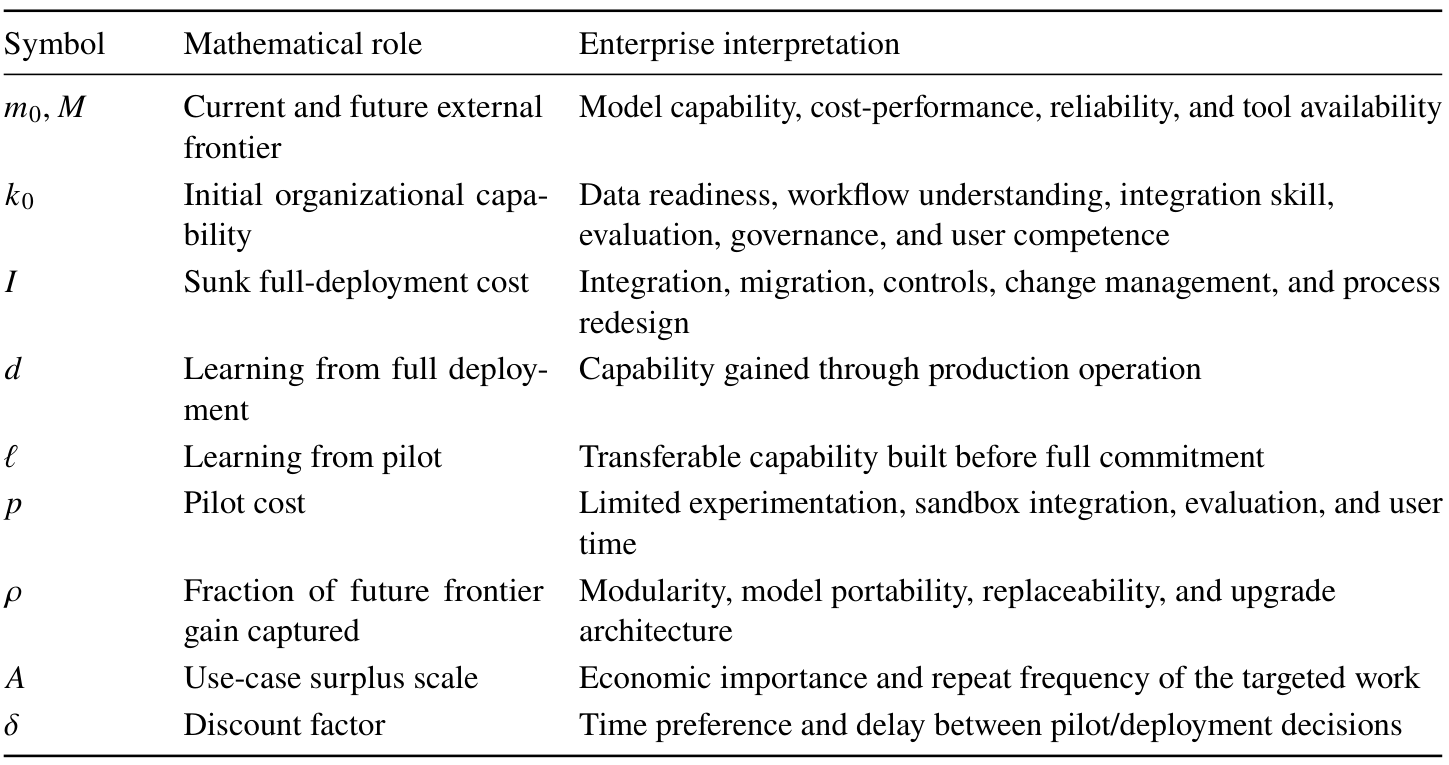}
\caption{Model notation and enterprise interpretation.}
\end{table}
\section{Analytical Results}
\subsection{Uncertainty favors options over affine commitment}
Proposition 1 (Uncertainty and the action set). Let \(M'\) be a mean-preserving spread of \(M\). Then

\begin{equation}
W(M')\ge W(M),\qquad P(M')\ge P(M),\qquad D(M')=D(M).
\end{equation}
The inequalities are strict when the additional dispersion places positive probability on both sides of the relevant deployment threshold.

The result follows from convexity. The date-1 option payoff \(h_k(m)\) is the maximum of an affine payoff and zero. Greater dispersion raises the expectation of a convex payoff. Immediate deployment is different because Equation (3) is affine in \(M\) and depends only on \(\mu_M\). The proposition clarifies a frequent ambiguity. ``Uncertainty slows adoption'' is incomplete. Uncertainty raises the value of deferring an irreversible deployment, but it can also raise the value of a pilot whose downside is bounded and whose capability is useful in high-frontier states. Whether the firm becomes passive or experimental depends on the pilot's learning economics.

\subsection{The AI waiting paradox}
Consider a location shift \(M_x=M+x\), where \(x\) represents a uniform improvement in the future frontier. Assume \(M\) has a continuous distribution. Differentiating the option values gives the next result.

Proposition 2 (Faster progress can delay commitment). For a location shift in the future frontier,

\begin{equation}
\frac{\partial W}{\partial x}=\delta A k_0\,\mathbb{P}\!\left(Ak_0(M+x)>I\right).
\end{equation}
\begin{equation}
\frac{\partial P}{\partial x}=\delta A(k_0+\ell)\,\mathbb{P}\!\left(A(k_0+\ell)(M+x)>I\right).
\end{equation}
\begin{equation}
\frac{\partial D}{\partial x}=\delta A(k_0+d)\rho.
\end{equation}
Therefore, faster expected frontier progress raises waiting relative to immediate deployment whenever

\begin{equation}
k_0\,\mathbb{P}\!\left(Ak_0(M+x)>I\right)>\rho(k_0+d).
\end{equation}
and raises piloting relative to deployment whenever

\begin{equation}
(k_0+\ell)\,\mathbb{P}\!\left(A(k_0+\ell)(M+x)>I\right)>\rho(k_0+d).
\end{equation}
These inequalities define the AI waiting paradox. A better expected future frontier can make AI economically more attractive while making today's full deployment relatively less attractive. The paradox is most likely when the installed architecture has low modularity, the future adoption option is likely to be in the money, and a pilot creates substantial transferable capability. The result is not an argument for delay in general. When \(\rho\) is high, immediate deployment captures much of the external improvement. When \(d\) is large, production learning can outweigh the option value of waiting. And when the current operating payoff \(Ak_0m_0\) is large, deployment receives a benefit that neither waiting nor piloting receives.

\subsection{When a pilot is economically justified}
Define the incremental date-1 payoff created by pilot learning:

\begin{equation}
\Delta(m;\ell)=h_{k_0+\ell}(m)-h_{k_0}(m).
\end{equation}
Then

\begin{equation}
P-W=-p+\delta\,\mathbb{E}[\Delta(M;\ell)].
\end{equation}
Proposition 3 (Pilot threshold). A pilot weakly dominates pure waiting if and only if

\begin{equation}
p\le p^*(\ell)\equiv\delta\,\mathbb{E}[\Delta(M;\ell)].
\end{equation}
The threshold \(p^*(\ell)\) is nondecreasing in \(\ell\) and, almost everywhere,

\begin{equation}
\frac{\partial p^*}{\partial\ell}=\delta A\,\mathbb{E}\!\left[M\,\mathbf{1}\{A(k_0+\ell)M>I\}\right]>0.
\end{equation}
whenever the pilot has a positive probability of enabling profitable deployment. Moreover, \(p^*(\ell)\) weakly increases under any first-order stochastic improvement in \(M\).

The key object is not the pilot's activity level, user count, or prototype quality. It is the expected increase in the value of the later deployment decision. A pilot that produces a polished demonstration but no transferable capability has \(\ell\) close to zero and cannot justify much cost. A pilot that teaches the firm how to map the workflow, evaluate failure, structure data access, and integrate a replaceable model can have a larger \(\ell\) even if its immediate operating impact is modest. The final statement follows because \(\Delta(m;\ell)\) is nondecreasing in \(m\). Better future technology makes organization-specific capability more valuable, so the willingness to pay for learning rises with the frontier. This result separates two effects of rapid progress: it may reduce the attractiveness of locking in today's architecture while increasing the value of building the capability to exploit tomorrow's models.

\subsection{A nonempty pilot-early, commit-late region}
Proposition 4 (Existence of staged adoption). Suppose \(M\ge m_0>0\) almost surely and \(\mathbb{E}[M]<\infty\). For any fixed finite \(p\), \(D(\rho)\), and \(W\), there exists \(\bar{\ell}\) < \(\infty\) such that for all \(\ell>\bar{\ell}\),

\begin{equation}
P>\max\{D(\rho),W\}.
\end{equation}
More generally, for any finite \(\ell\), piloting is optimal exactly when

\begin{equation}
p<\delta\,\mathbb{E}[h_{k_0+\ell}(M)]-\max\{D(\rho),W\}.
\end{equation}
Piloting occupies a nonempty region of the parameter space. As transferable learning rises, date-1 deployment becomes profitable in more states while the pilot retains the option not to deploy. A pilot is not optimal because it is smaller or politically easier. It is optimal only if the capability it creates is sufficiently valuable relative to its cost and to the current operating value forgone by not deploying fully.

\subsection{The modularity threshold}
Write the deployment value as

\begin{equation}
\begin{aligned}
D(\rho)&=D_0+B\rho,\qquad D_0=Ak_0m_0-I+\delta A(k_0+d)m_0,\\
B&=\delta A(k_0+d)(\mu_M-m_0)>0.
\end{aligned}
\end{equation}
Let \(V_o=\max\{P,W\}\) denote the best outside option.

Proposition 5 (Critical modularity). The raw modularity threshold at which deployment ties the best outside option is

\begin{equation}
\rho^*=\frac{V_o-D_0}{\delta A(k_0+d)(\mu_M-m_0)}.
\end{equation}
If \(0<\rho^*<1\), immediate deployment is optimal if and only if \(\rho\ge\rho^*\). If \(\rho^*\le0\), deployment dominates for every feasible \(\rho\). If \(\rho^*\) > 1, no feasible modularity level makes immediate deployment optimal.

The threshold converts an architectural property into an investment rule. A system can justify early deployment either by delivering high current operating surplus or by retaining enough flexibility to absorb future model improvement. Low current value requires more modularity; a valuable pilot or waiting option also raises the hurdle. The threshold need not fall monotonically with expected frontier progress. At first, a higher \(\mu_M\) can make a modular deployment more attractive by enlarging the improvement it captures. At higher levels, however, the pilot option can rise faster than deployment, causing \(\rho^*\) to increase. The numerical section shows this non-monotonic pattern under an illustrative parameterization. It is a conditional result, not a universal theorem.

\subsection{Where the learning occurs}
Proposition 6 (Learning-channel substitution). The marginal values of learning from deployment and from piloting are

\begin{equation}
\frac{\partial D}{\partial d}=\delta A\left[m_0+\rho(\mu_M-m_0)\right]>0.
\end{equation}
\begin{equation}
\frac{\partial P}{\partial\ell}=\delta A\,\mathbb{E}\!\left[M\,\mathbf{1}\{A(k_0+\ell)M>I\}\right]\ge0.
\end{equation}
Thus, production learning shifts the firm toward deployment, whereas pilot-specific learning shifts it toward staged adoption.

The result identifies the practical question hidden inside ``learn by doing'': doing what? If the capabilities learned in production are inseparable from a brittle architecture, \(d\) may be large while \(\rho\) remains low. If a pilot is designed around portable evaluation, workflow mapping, and model interchangeability, \(\ell\) can be valuable even without production scale. The model therefore rejects the idea that every early deployment is automatically strategic learning.

\subsection{Competitive preemption as a reduced-form extension}
Competition can erode the option to wait. Let \(c_W\ge0\) and \(c_P\ge0\) be the opportunity costs of deferring market entry under waiting and piloting. Replacing \(W\) with \(W-c_W\) and \(P\) with \(P-c_P\) leaves the propositions intact but lowers the outside option \(V_o\) and therefore lowers \(\rho^*\). The firm deploys earlier when delaying sacrifices customer access, data rights, distribution, standards influence, or scarce implementation capacity. This extension is intentionally reduced form; strategic interaction among multiple adopters is left for future work.

\newpage
\section{Numerical Experiments}
The analytical results do not require a distributional assumption. To visualize policy regions, let

\begin{equation}
M=m_0+Y,\qquad Y\sim\operatorname{Lognormal}(\nu,\sigma^2),\qquad \mathbb{E}[M]=\mu_M.
\end{equation}
The shifted lognormal ensures \(M\ge m_0\) while permitting right-skewed frontier outcomes. The expected option payoff has a closed form reported in Appendix B. All figures are generated from the analytical formula, not from Monte Carlo simulation. The baseline normalization is

\begin{equation}
A=1,\quad k_0=0.8,\quad I=2,\quad \delta=0.95,\quad m_0=1,\quad d=0.2,\quad \ell=0.4,\quad p=0.12.
\end{equation}
These values are illustrative. They are chosen to display economically distinct policy regions and should not be interpreted as estimated enterprise returns or costs.

\subsection{Three policy regions}
Figure 2 plots the optimal action as expected future capability and modularity vary at \(\sigma=0.45\). When expected frontier progress is low, waiting dominates because neither the current deployment nor the pilot has enough value to cover its cost. At higher expected progress and low modularity, piloting dominates: tomorrow's frontier is valuable, but today's installed architecture would capture too little of it. At high modularity, immediate deployment dominates because the firm receives current operating value while retaining exposure to future improvement. Table 3 reports representative points. At \(\mu_M=1.8\) and \(\rho=0.2\), the pilot value is 0.117, versus 0.014 for waiting and -0.098 for deployment. Raising modularity to 0.8 leaves the two option values unchanged but raises deployment to 0.358. At \(\mu_M=2.4\), low modularity makes the pilot strongly dominant; high modularity makes deployment dominant. The same expected frontier can therefore support opposite actions depending on architecture.

\begin{figure}[H]
\centering
\includegraphics[width=0.92\textwidth]{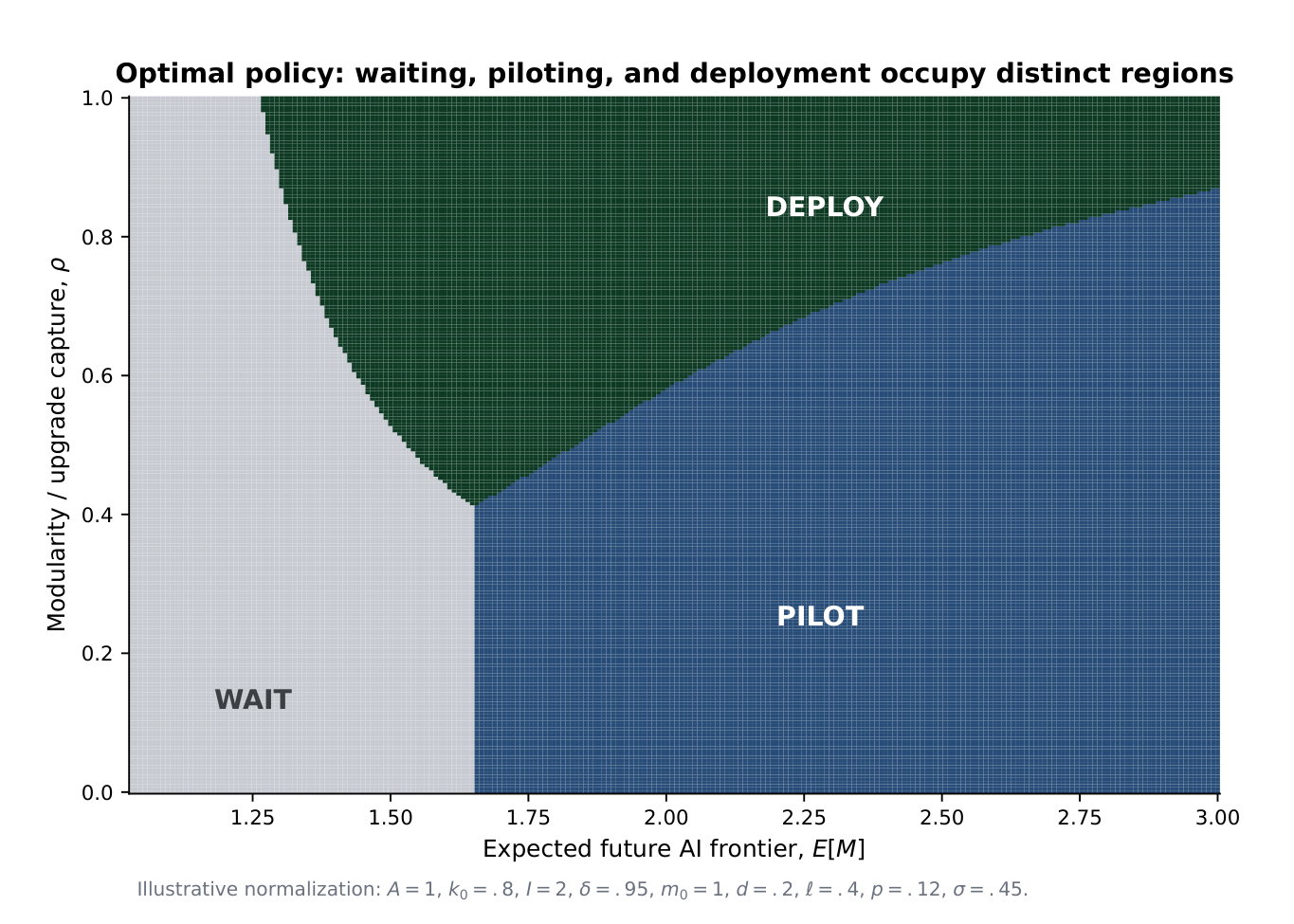}
\caption{Optimal policy under the baseline normalization. The existence of three regions illustrates that pilot and wait are economically distinct responses to rapid progress.}
\end{figure}

\begin{table}[H]
\centering
\includegraphics[width=0.88\textwidth]{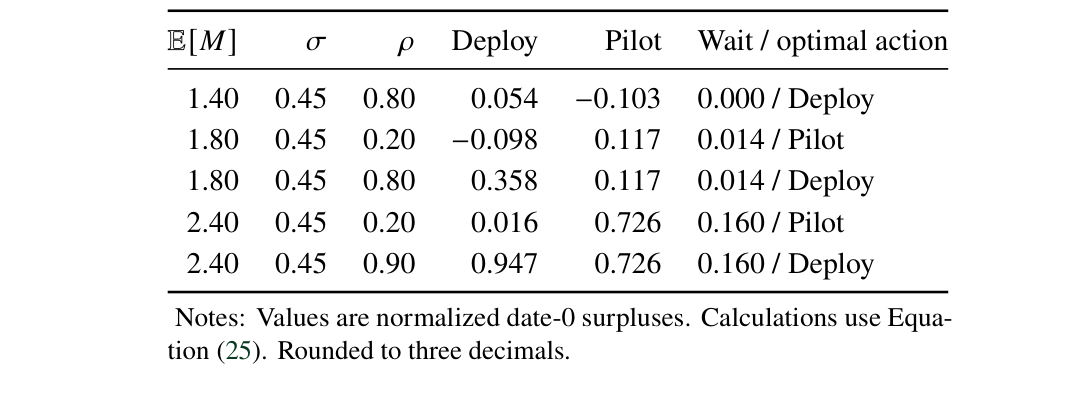}
\caption{Illustrative action values under the baseline normalization.}
\end{table}

\subsection{The economics of a useful pilot}
Figure 3 varies pilot learning \(\ell\) and cost \(p\) while holding \(\mu_M=2.05\), \(\sigma=0.55\), and \(\rho=0\). The boundary is exactly the threshold in Equation (16). The shape is nonlinear because learning first increases the probability that later deployment becomes profitable and then increases the surplus conditional on deployment. Equation (16) gives a falsifiable standard for pilot design. If a proposed pilot cannot identify which transferable capability it creates, its implied \(\ell\) is not established. If the capability is specific to a model or temporary interface, it may not survive to the date-1 deployment decision. A pilot can therefore be active and expensive while having low option value.

\begin{figure}[H]
\centering
\includegraphics[width=0.92\textwidth]{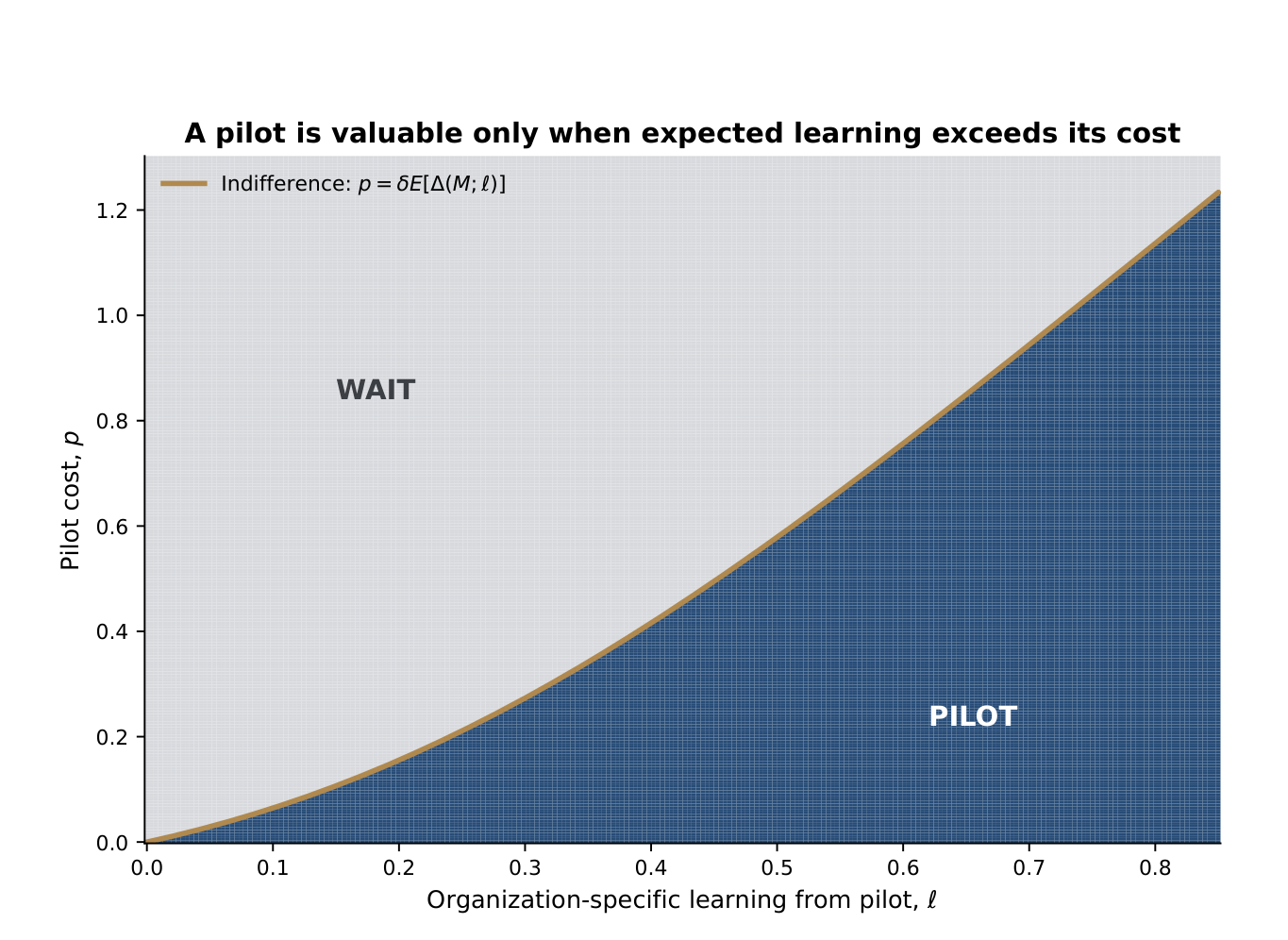}
\caption{Pilot versus wait. The gold curve is the exact indifference condition. Piloting is optimal below the curve; passive waiting is optimal above it.}
\end{figure}

\subsection{Uncertainty and the ranking of actions}
Figure 4 holds the mean frontier fixed at 1.8 and varies \(\sigma\). Deploy value is horizontal because it is affine in \(M\). Pilot and wait values rise with uncertainty. At low modularity, deployment is already unattractive and the pilot becomes increasingly valuable. At high modularity, deployment remains preferred over the plotted range, but its advantage narrows. The figure does not imply that firms should seek uncertainty. It shows that uncertainty changes which organizational actions are valuable. A small pilot can be attractive in volatile environments precisely because it combines limited downside with capability that pays off in high-frontier states.

\begin{figure}[H]
\centering
\includegraphics[width=0.92\textwidth]{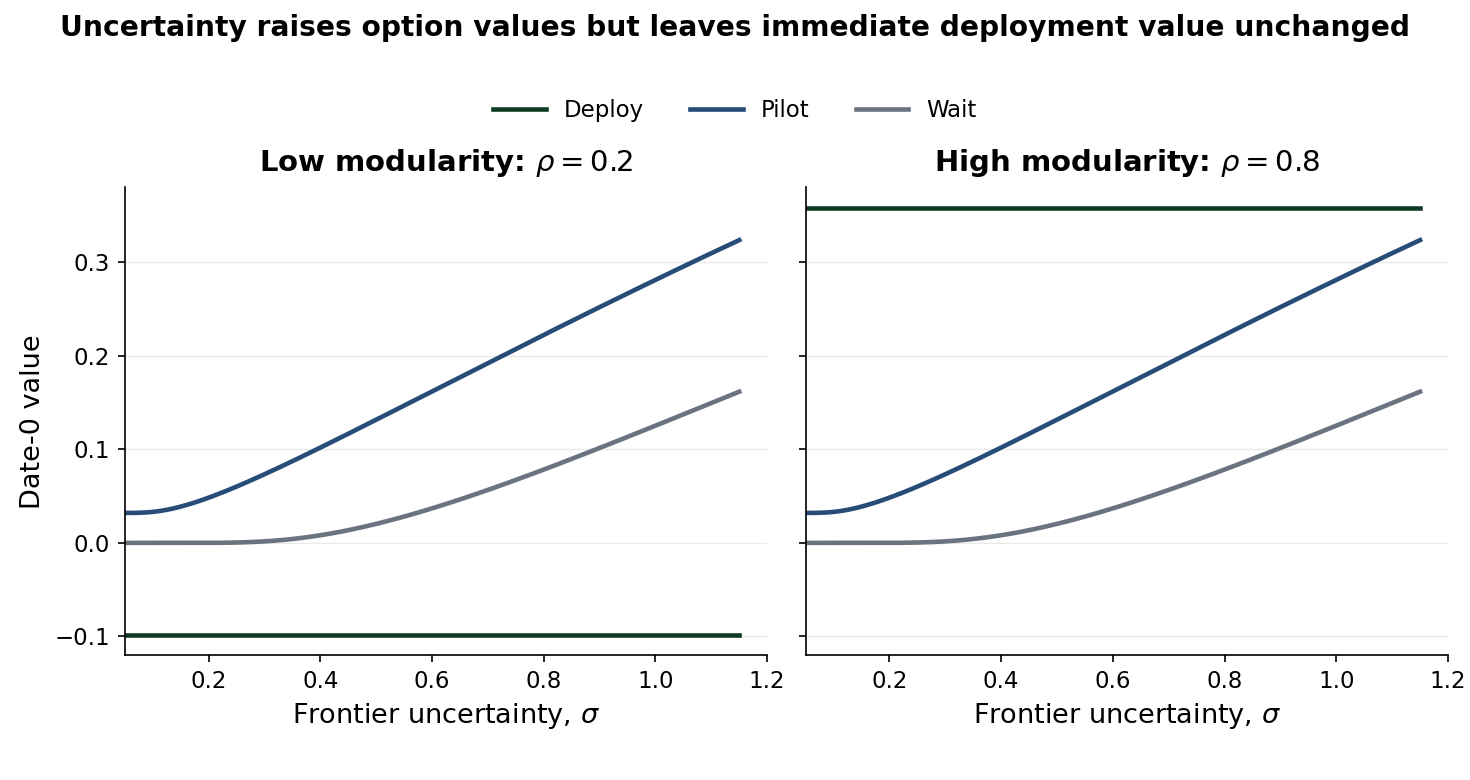}
\caption{Mean-preserving frontier uncertainty. Option values rise because the firm can decline to deploy in low states while retaining upside in high states. Immediate deployment value is unchanged under the model's affine payoff.}
\end{figure}

\subsection{The modularity hurdle}
Figure 5 plots the raw threshold in Equation (21). Values above one mean that no feasible architecture in the normalized model makes immediate deployment optimal. Within the feasible band, deployment is optimal above the curve. The U-shape has a direct interpretation. Moderate expected progress strengthens the case for a modular deployment. Very strong expected progress also strengthens the pilot, because a capability-building experiment provides access to a larger future upside without locking in the present architecture. More technical progress can therefore increase, rather than reduce, the amount of modularity required to justify commitment.

\begin{figure}[H]
\centering
\includegraphics[width=0.92\textwidth]{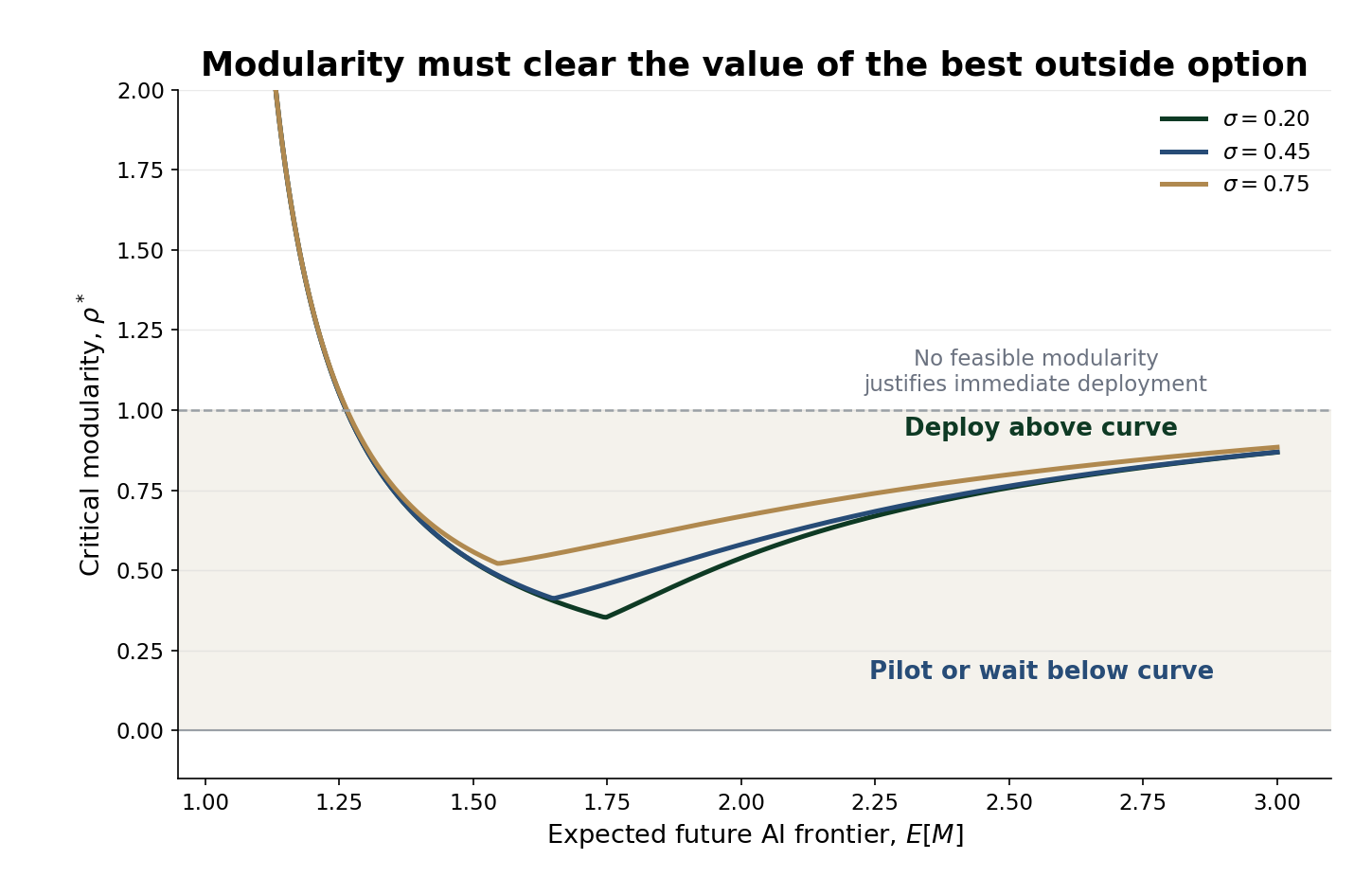}
\caption{Critical modularity under three uncertainty levels. The threshold initially falls as a more valuable frontier makes upgrade capture useful, then rises when the pilot option grows faster than immediate deployment. The non-monotonicity is numerical, not asserted as universal.}
\end{figure}

\section{Continuous-Time Extension}
The two-period model makes the three-way action set transparent. A standard continuous-time extension confirms the core timing effects. Let the external AI frontier follow geometric Brownian motion

\begin{equation}
dX_t=\mu X_t\,dt+\sigma X_t\,dB_t,\qquad r>\mu.
\end{equation}
where \(r\) is the discount rate. Once deployed, a firm with capability \(k\) receives perpetual cash flow \(q(\rho)kX_t\). Modularity raises frontier capture according to

\begin{equation}
q(\rho)=q_0(1+\alpha\rho),\qquad \alpha\ge0.
\end{equation}
and reduces the effective sunk portion of deployment cost according to

\begin{equation}
I(\rho)=I_0(1-\gamma\rho),\qquad 0\le\gamma<1.
\end{equation}
The project value upon adoption is

\begin{equation}
J(X;\rho,k)=\frac{q(\rho)kX}{r-\mu}-I(\rho).
\end{equation}
Let \(\beta>1\) be the positive root of

\begin{equation}
\frac{1}{2}\sigma^2\beta(\beta-1)+\mu\beta-r=0.
\end{equation}
The standard value-matching and smooth-pasting conditions yield the adoption threshold

\begin{equation}
X^*(\rho,k,\sigma)=\frac{\beta(r-\mu)I_0(1-\gamma\rho)}{(\beta-1)q_0(1+\alpha\rho)k}.
\end{equation}
Corollary 1 (Continuous-time comparative statics). Under the stated restrictions,

\begin{equation}
\frac{\partial X^*}{\partial\sigma}>0,\qquad
\frac{\partial X^*}{\partial k}<0,\qquad
\frac{\partial X^*}{\partial\rho}<0.
\end{equation}
Uncertainty raises the hurdle at which the firm exercises the irreversible adoption option. Organizational capability and modularity lower it. Figure 6 illustrates the result for normalized parameters. A high-capability, highly modular organization can rationally deploy at a much lower frontier level than a low-capability organization facing the same technical environment. The extension omits piloting and asks only whether the discrete model's comparative statics for uncertainty, capability, and modularity survive in the canonical continuous-time setting. Repeated pilot stages are left to a richer dynamic model.

\begin{figure}[H]
\centering
\includegraphics[width=0.92\textwidth]{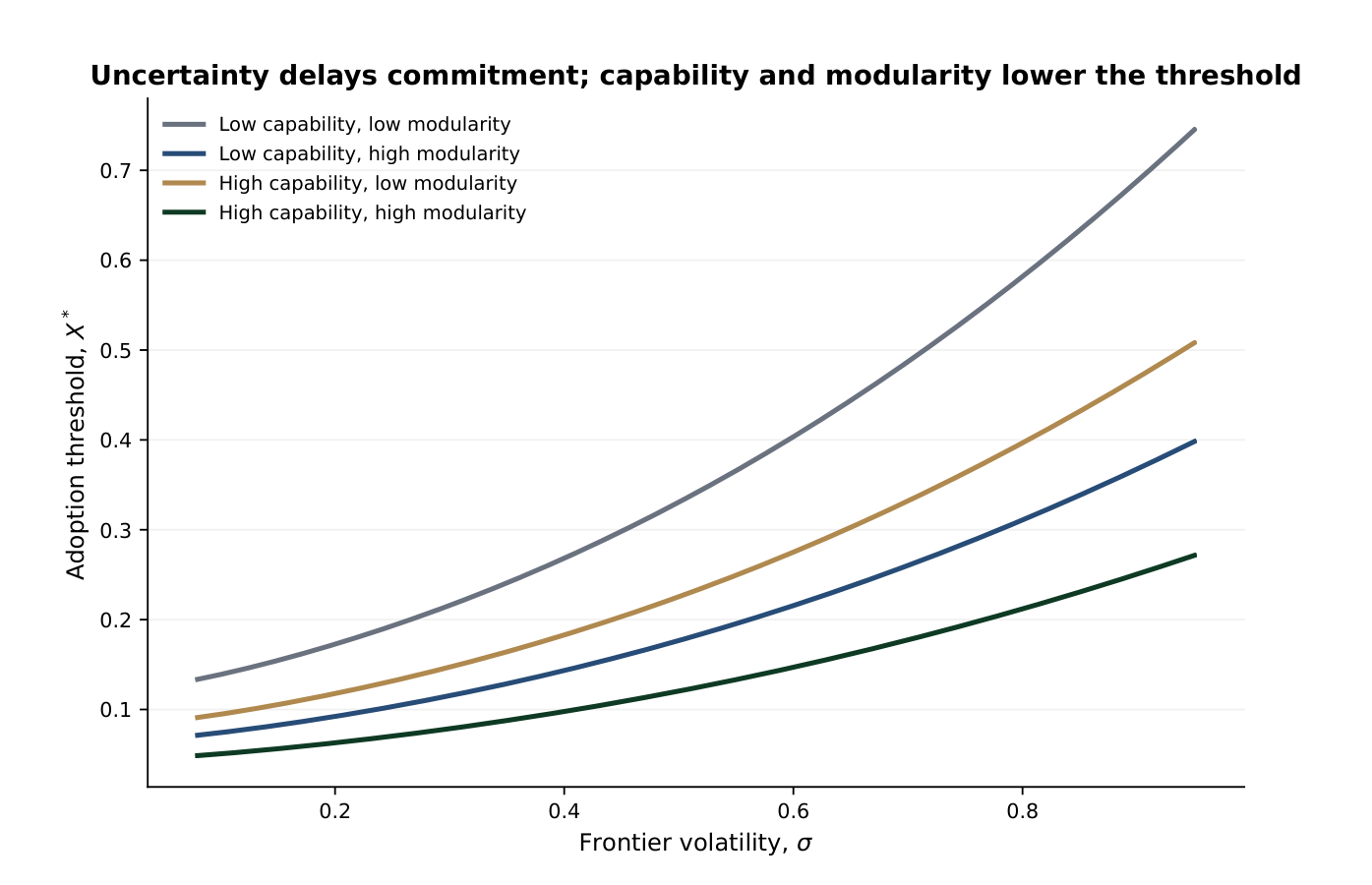}
\caption{Continuous-time adoption threshold. The numerical curves use $r=0.10$, $\mu=0.025$, $I_0=q_0=1$, $\gamma=0.35$, and $\alpha=0.60$.}
\end{figure}

\section{Implications}
\subsection{The speed of adoption is not the same as the quality of timing}
The model implies no general rule to deploy faster or to wait longer. A fast-moving frontier increases the value of AI, but not necessarily the value of today's irreversible architecture. Early deployment is rational when current operating surplus is large, production learning is valuable, and the system can absorb future progress. Waiting is rational when the option value is high and a pilot would create little transferable capability. Piloting is rational when bounded experimentation creates capability that survives the model cycle. This distinction matters for interpreting adoption statistics. A firm with many disconnected pilots may be failing to commit, or it may be building valuable capability. A firm with a large production deployment may be ahead, or it may have exercised too early into an architecture with low \(\rho\). Deployment status alone cannot identify which state the firm occupies.
\subsection{Pilot design should maximize transferable learning}
Equation (16) provides the central design rule: maximize \(\mathbb{E}[\Delta(M;\ell)]\) per dollar of pilot cost. That favors pilots that answer questions likely to matter after the underlying model changes. Examples include whether the workflow has a stable economic owner, whether source data can be accessed lawfully and reliably, which exceptions require judgment, how output quality should be measured, where system boundaries sit, and whether users alter their behavior when the tool is introduced. A pilot built around a temporary benchmark or a single model's current interface may demonstrate capability without creating much \(\ell\). A pilot built around portable evaluation, documented interfaces, workflow instrumentation, and a model-independent control layer can create capability that remains useful when the frontier moves.

\subsection{Modularity is an economic hedge against model progress}
The usual case for modularity is engineering flexibility. Proposition 5 adds an investment interpretation. Modularity raises the share of future frontier gains captured by an early deployment and can reduce the effective sunk cost of changing models, vendors, or components. It therefore lowers the option premium required to justify commitment. This does not mean every system should maximize abstraction. Modularity can increase current cost, latency, complexity, and coordination burden. The model treats \(\rho\) as a benefit without separately pricing those costs. The practical decision is whether the incremental architecture cost is less than the increase in deployment value produced by a higher \(\rho\).

\subsection{Enterprise deployment evaluation}
The framework suggests four questions for evaluating enterprise AI deployments. 1. Current surplus: What measurable operating value is available from the present frontier, before crediting hypothetical future capability? 2. Learning transfer: What capability will the pilot create that remains useful if the model, vendor, or interface changes? 3. Upgrade capture: What fraction of frontier improvement can the deployed architecture absorb without repeating the implementation cost? 4. Exercise discipline: What observable condition converts the pilot into deployment, and what condition causes abandonment? These questions separate a valuable option from indefinite experimentation. A pilot without an exercise rule can become a recurring cost center. A deployment without an architecture for future capture can become premature lock-in.

\subsection{Policy and ecosystem design}
The model also has a policy implication. Subsidizing access to models does not necessarily accelerate productive deployment if the binding constraint is organization-specific capability. Shared evaluation resources, interoperable standards, implementation templates, and workforce training can raise \(k_0\) or \(\ell\). Procurement and technical standards that preserve model portability can raise \(\rho\). Those interventions affect the adoption threshold differently from a subsidy that simply lowers token prices.

\section{Limitations and Empirical Agenda}
The model is intentionally sparse and should be interpreted accordingly.

Two periods. Enterprise AI programs unfold through repeated pilots, upgrades, and abandonment decisions. The two-period structure isolates the first commitment decision but cannot represent path dependence across many model generations.

One-dimensional frontier. Model capability is jagged across tasks (Dell'Acqua et al., 2026). A scalar \(M\) compresses quality, cost, latency, reliability, security, and modality into one payoff-relevant index. A multidimensional frontier could generate portfolio choices across workflows rather than a single adoption decision.

Deterministic learning increments. The model treats \(d\) and \(\ell\) as known. In reality, a pilot produces uncertain learning and may reveal whether the use case is viable. Adding Bayesian signals would make experimentation valuable through both capability accumulation and information.

Risk neutrality and internal finance. The firm is risk neutral and can fund every action. Financing constraints, managerial incentives, and accounting treatment may alter the ranking of pilots and deployment.

No strategic competition. Competitors, customers, regulators, and vendors can reduce the value of waiting. The reduced-form preemption costs in Section 4.6 do not capture equilibrium adoption races, standards competition, or supplier bargaining.

No architecture cost function. Modularity is beneficial in the model but not free. An empirical application should estimate the current cost of portability and the probability that it preserves future value. The model predicts that higher frontier uncertainty should reduce irreversible deployment but can increase capability-building pilots; that effect should be strongest when pilots create portable learning. The relative deployment response to model progress should be more positive for systems with model abstraction, standardized interfaces, and replaceable evaluation layers. Pilot-to-production conversion should be higher when pilot artifacts transfer directly into production rather than being rebuilt. Finally, organizations with stronger absorptive capacity should have lower deployment thresholds even after controlling for use-case economics. A credible empirical design would require project-level longitudinal data with explicit stage transitions, architecture characteristics, pilot objectives, and realized operating outcomes. Public procurement inventories, software repositories, cloud architecture records, and repeated enterprise surveys may provide partial routes. The theory is designed to state what such data should test, not to substitute a stylized numerical exercise for empirical identification.

\section{Conclusion}
Rapid AI progress creates a strategic temptation to collapse three different actions into one imperative: move faster. The model shows why that is incomplete. A firm can move by deploying, by piloting, or by preserving the option to wait. Each action has a different exposure to uncertainty, learning, and obsolescence. The central result is a dual response to the moving frontier. Greater uncertainty increases the value of waiting because downside can be avoided. It also increases the value of a pilot when organization-specific learning creates more upside in high-frontier states. Faster expected progress can delay full deployment when the installed architecture captures too little of that progress. In that region, the rational policy is not passivity. It is to pilot early and commit late. The economics of a pilot therefore depend on what survives it. A temporary demonstration has little option value if it creates no transferable capability. A well-designed pilot creates workflow knowledge, evaluation infrastructure, integration competence, and governance capacity that remain useful when the model changes. Modularity performs the complementary role on the deployment side: it allows an early commitment to capture more of the future frontier. For enterprise AI, the relevant strategic question is not simply whether to adopt. It is which commitment should be irreversible, which learning should be purchased now, and which parts of the system must remain replaceable. Firms that answer those questions well can act early without betting the organization on today's model.

\section*{Disclosure}
The author is affiliated with Omega Venture Partners, an investment firm focused on artificial intelligence and enterprise software, and may hold investments whose value could be affected by the themes discussed. This is a theoretical paper. Its numerical experiments are illustrative normalizations, not empirical estimates, forecasts, or investment recommendations. The paper uses no confidential company, portfolio, customer, founder, diligence, or transaction information. No external funding was received. Generative AI tools assisted with language editing, typesetting, and implementation of numerical checks. The author verified all analytical and numerical results and the final text.

\section{Data and Code Availability}
The paper contains no proprietary or confidential data. All numerical figures are generated from the analytical formulas in the model using the illustrative parameter values reported in the text. Appendix B gives the closed-form expectation used in the numerical exercises, and Appendix C states the numerical verification protocol. No separate data or computational package accompanies this version.

\appendix
\section{Proofs}
Proof of Proposition 1. For any $k>0$, $h_k(m)=\max\{Akm-I,0\}$ is convex because it is the pointwise maximum of two affine functions. If $M'$ is a mean-preserving spread of $M$, the definition of convex order implies $\mathbb{E}[h_k(M')]\ge\mathbb{E}[h_k(M)]$. Applying this result at $k=k_0$ and $k=k_0+\ell$ proves the inequalities for $W$ and $P$. Deployment value in Equation (3) depends on $M$ only through $\mathbb{E}[M]$, which is unchanged by a mean-preserving spread. Strictness obtains when the spread changes mass across the kink $I/(Ak)$; otherwise the payoff is affine over the relevant support and equality can hold. $\square$

Proof of Proposition 2. Let $g(x)=\mathbb{E}[[b(M+x)-I]_+]$ for $b>0$. Because $M$ has a continuous distribution, the payoff is differentiable in $x$ almost surely and its derivative is $b\mathbf{1}\{b(M+x)>I\}$. The derivative is bounded by $b$, so dominated convergence permits differentiation under the expectation:
\[
g'(x)=b\,\mathbb{P}\!\left(b(M+x)>I\right).
\]
Substituting $b=Ak_0$ and $b=A(k_0+\ell)$ and multiplying by $\delta$ yields the first two derivatives. Equation (3) gives the third derivative directly. Subtracting the deployment derivative from the other two gives inequalities (12) and (13). $\square$

Proof of Proposition 3. Equation (15) follows by subtracting $W$ from $P$, so $P\ge W$ if and only if Equation (16) holds. The function $h_{k_0+\ell}(m)$ is nondecreasing in $\ell$ for every $m>0$, hence $p^*(\ell)$ is nondecreasing. At values of $\ell$ for which the threshold event has zero probability, differentiation under the expectation gives
\[
\frac{\partial}{\partial\ell}h_{k_0+\ell}(M)=AM\,\mathbf{1}\{A(k_0+\ell)M>I\},
\]
which yields Equation (17). To establish the stochastic-order result, note that $\Delta(m;\ell)$ is piecewise linear and nondecreasing in $m$: it is zero below $I/[A(k_0+\ell)]$, rises with slope $A(k_0+\ell)$ until $I/(Ak_0)$, and then rises with slope $A\ell$. The expectation of a nondecreasing function weakly rises under first-order stochastic dominance. $\square$

Proof of Proposition 4. Because $M\ge m_0>0$, choose $\ell$ large enough that $A(k_0+\ell)m_0>I$. Then $h_{k_0+\ell}(M)=A(k_0+\ell)M-I$ in every state and
\[
P=-p+\delta\left[A(k_0+\ell)\mathbb{E}[M]-I\right].
\]
This expression is affine and unbounded in $\ell$, while $D(\rho)$ and $W$ do not depend on $\ell$. Therefore a finite $\bar{\ell}$ exists above which $P$ strictly dominates both. For fixed $\ell$, rearranging $P>\max\{D(\rho),W\}$ yields Equation (19). $\square$

Proof of Proposition 5. Equation (20) follows by collecting the terms in $\rho$ in Equation (3). Because $\mu_M>m_0$, $B>0$, so $D(\rho)$ is strictly increasing and crosses the constant outside option $V_o$ at most once. Solving $D_0+B\rho=V_o$ gives Equation (21). The three cases follow from whether the crossing lies below, inside, or above the feasible interval $[0,1]$. $\square$

Proof of Proposition 6. The derivative of Equation (3) with respect to $d$ is immediate. The derivative of Equation (6) with respect to $\ell$ is the same dominated-convergence calculation used in Proposition 3. Both derivatives are nonnegative and are strictly positive under the conditions stated. $\square$

Proof of Corollary 1. The standard optimal-stopping solution for a perpetual project with geometric Brownian state variable is $F(X)=CX^\beta$ below the adoption boundary, where $\beta>1$ solves Equation (30). Value matching and smooth pasting against $J(X;\rho,k)$ yield Equation (31). The derivatives with respect to $k$ and $\rho$ are negative by inspection. For volatility, implicit differentiation of Equation (30) gives
\[
\frac{d\beta}{d\sigma}=-\frac{\sigma\beta(\beta-1)}{\tfrac{1}{2}\sigma^2(2\beta-1)+\mu}<0,
\]
where the denominator is the positive derivative of the quadratic at its larger root. Since $d[\beta/(\beta-1)]/d\beta=-1/(\beta-1)^2<0$, the threshold factor $\beta/(\beta-1)$ increases with $\sigma$. $\square$

\section{Closed-Form Numerical Expectation}
Let $Y\sim\operatorname{Lognormal}(\nu,\sigma^2)$ and $M=m_0+Y$. The mean restriction is
\begin{equation}
\exp\!\left(\nu+\frac{1}{2}\sigma^2\right)=\mu_M-m_0.
\end{equation}
For a slope $b>0$, define $y^*=I/b-m_0$. If $y^*\leq 0$, the option is in the money in every state and
\begin{equation}
\mathbb{E}\!\left[[bM-I]_+\right]=b\mu_M-I.
\end{equation}
If $y^*>0$, define
\begin{equation}
z_0=\frac{\nu-\ln y^*}{\sigma},\qquad z_1=z_0+\sigma.
\end{equation}
Using the truncated first moment of a lognormal variable,
\begin{equation}
\mathbb{E}\!\left[[bM-I]_+\right]
=b\left[m_0\Phi(z_0)+(\mu_M-m_0)\Phi(z_1)\right]-I\Phi(z_0),
\end{equation}
where $\Phi$ is the standard normal distribution function. The implementation was checked against independent numerical quadrature across 48 parameter combinations; the maximum absolute discrepancy was below $10^{-14}$.

\section{Numerical Verification Protocol}
The numerical audit applies the following checks after figure generation:
\begin{enumerate}[leftmargin=*,itemsep=0.15em,topsep=0.25em]
\item Compare Equation (36) with independent adaptive quadrature across 48 combinations of slopes, means, and volatility levels.
\item Verify that mean-preserving increases in the lognormal dispersion weakly raise \(W\) and \(P\) while leaving \(D\) unchanged.
\item Compare the analytical derivatives in Proposition 2 with common-random-number finite differences using two million draws.
\item Verify monotonicity of the pilot threshold in \(\ell\) and the identity \(P-W=p^*(\ell)-p\) around the pilot--wait boundary.
\item Verify that Equation (21) produces an exact tie and the correct policy switch on both sides of the threshold.
\item Verify that all three policy regions are nonempty under the baseline normalization.
\item Verify the continuous-time threshold comparative statics.
\item Scan a broad parameter grid for nonfinite outputs.
\end{enumerate}
Independent recomputation of the benchmark action values and continuous-time comparative statics is used as a cross-check. Numerical validation supports the stated formulas; it does not turn the illustrative normalization into an empirical calibration.

\section*{References}
Adner, R. and Levinthal, D. A. (2004). What is not a real option: Considering boundaries for the application of real options to business strategy. Academy of Management Review, 29(1):74--85.\par\medskip
Banerjee, B. and Singh, S. (2026). The Human--AI Substitution Principle: When will you be replaced by AI in your organization? arXiv:2607.20781.\par\medskip
Benaroch, M. and Kauffman, R. J. (1999). A case for using real options pricing analysis to evaluate information technology project investments. Information Systems Research, 10(1):70--86.\par\medskip
Bloom, N., Bond, S., and Van Reenen, J. (2007). Uncertainty and investment dynamics. Review of Economic Studies, 74(2):391--415.\par\medskip
Bonney, K., Breaux, C. L., Dinlersoz, E., Foster, L. S., Haltiwanger, J. C., and Pande, A. A. (2026). The microstructure of AI diffusion: Evidence from firms, business functions, and worker tasks. Working Paper 35141, National Bureau of Economic Research.\par\medskip
Brynjolfsson, E., Li, D., and Raymond, L. (2025). Generative AI at work. Quarterly Journal of Economics, 140(2):889--942.\par\medskip
Brynjolfsson, E., Rock, D., and Syverson, C. (2021). The productivity J-curve: How intangibles complement general purpose technologies. American Economic Journal: Macroeconomics, 13(1):333--372.\par\medskip
Caplin, A. (2026). Planning capital and discovery-based learning-by-doing: Investment as staged discovery in the hyperscale era. Working Paper 35349, National Bureau of Economic Research.\par\medskip
Cohen, W. M. and Levinthal, D. A. (1990). Absorptive capacity: A new perspective on learning and innovation. Administrative Science Quarterly, 35(1):128--152.\par\medskip
Crouzet, N. and Eberly, J. C. (2026). R\&D uncertainty and cycles. Working Paper 34838, National Bureau of Economic Research.\par\medskip
Dell'Acqua, F., McFowland III, E., Mollick, E., Lifshitz, H., Kellogg, K. C., Rajendran, S., Krayer, L., Candelon, F., and Lakhani, K. R. (2026). Navigating the jagged technological frontier: Field experimental evidence of the effects of artificial intelligence on knowledge worker productivity and quality. Organization Science, 37(2):403--423.\par\medskip
Dixit, A. K. and Pindyck, R. S. (1994). Investment under Uncertainty. Princeton University Press, Princeton, NJ.\par\medskip
Erdogmus, H. (2002). Valuation of learning options in software development under private and market risk. The Engineering Economist, 47(3):308--353.\par\medskip
Fichman, R. G. (2004). Real options and IT platform adoption: Implications for theory and practice. Information Systems Research, 15(2):132--154.\par\medskip
Fridgen, G., Hartwich, E., Rägo, V., Rieger, A., and Stohr, A. (2022). Artificial intelligence as a call for retail banking: Applying digital options thinking to artificial intelligence adoption. In Proceedings of the 30th European Conference on Information Systems. Research Paper 103.\par\medskip
Gandhi, P., Kedia, S., Lu, J., Pan, J., and Wei, J. (2026). Awaiting the prompt? Generative AI uncertainty and firm-level investment. Working Paper 7038198, SSRN.\par\medskip
Grenadier, S. R. and Weiss, A. M. (1997). Investment in technological innovations: An option pricing approach. Journal of Financial Economics, 44(3):397--416.\par\medskip
Lee, C. P., Lee, M. K., and Mutlu, B. (2026). Making the invisible visible: Understanding the mismatch between organizational goals and worker experiences in AI adoption. arXiv:2605.03078.\par\medskip
McDonald, R. L. and Siegel, D. (1986). The value of waiting to invest. Quarterly Journal of Economics, 101(4):707--727.\par\medskip
McGrath, R. G. (1997). A real options logic for initiating technology positioning investments. Academy of Management Review, 22(4):974--996.\par\medskip
Nanda, R. and Rhodes-Kropf, M. (2016). Financing entrepreneurial experimentation. In Innovation Policy and the Economy, volume 16, pages 1--23. University of Chicago Press.\par\medskip
Pindyck, R. S. (1991). Irreversibility, uncertainty, and investment. Journal of Economic Literature, 29(3):1110--1148.\par\medskip
Ross, J.-M., Fisch, J. H., and Varga, E. (2018). Unlocking the value of real options: How firm-specific learning conditions affect R\&D investments under uncertainty. Strategic Entrepreneurship Journal, 12(3):335--353.\par\medskip
Schwartz, E. S. and Zozaya-Gorostiza, C. (2003). Investment under uncertainty in information technology: Acquisition and development projects. Management Science, 49(1):57--70.\par\medskip
Stanford Institute for Human-Centered Artificial Intelligence (2026). The 2026 AI index report. Stanford University.\par\medskip
Thomke, S. H. (1998). Managing experimentation in the design of new products. Management Science, 44(6):743--762.\par\medskip
Trigeorgis, L. and Reuer, J. J. (2017). Real options theory in strategic management. Strategic Management Journal, 38(1):42--63.\par\medskip
Yotzov, I., Barrero, J. M., Bloom, N., Bunn, P., Davis, S. J., Foster, K. M., Jalca, A., Meyer, B. H., Mizen, P., Navarrete, M. A., Smietanka, P., Thwaites, G., and Wang, B. Z. (2026). Firm data on AI. Working Paper 34836, National Bureau of Economic Research.\par\medskip
Yue, W. and Zhou, X. (2025). The option value of waiting: Firm valuation of AI investments under uncertainty. Working Paper 5954931, SSRN.\par\medskip
\end{document}